# Bridging Deep Learning and Integer Linear Programming: A Predictive-to-Prescriptive Framework for Supply Chain Analytics

Khai Banh Nghiep[1], Duc Nguyen Minh[2], Lan Hoang Thi[3]

[1] Tran Dai Nghia Secondary and High School, Vietnam  Email: Nghiepkhaibanh@gmail.com

[2] AI engineer Vietnam  Email: phucanhthien@gmail.com

[3] University of Languages and International studies  Email: Hoanglan1999@gmail.com

**Abstract:** *Although demand forecasting is a critical component of supply chain planning, actual retail data can exhibit irreconcilable seasonality, irregular spikes, and noise, rendering precise projections nearly unattainable. This paper proposes a three-step analytical framework that combines forecasting and operational analytics. The first stage consists of exploratory data analysis, where delivery-tracked data from 180,519 transactions are partitioned, and long-term trends, seasonality, and delivery-related attributes are examined. Secondly, the forecasting performance of a statistical time series decomposition model N-BEATS MSTL and a recent deep learning architecture N-HiTS were compared. N-BEATS and N-HiTS were both statistically, and hence were N-BEATS's and N-HiTS's statistically selected. Most recent time series deep learning models, N-HiTS, N-BEATS. N-HiTS and N-BEATS N-HiTS and N-HiTS outperformed the statistical benchmark to a large extent. N-BEATS was selected to be the most optimized model, as the one with the lowest forecasting error, in the 3rd and final stage forecasting values of the next 4 weeks of 1918 units, and provided those as a model with a set of deterministically integer linear program outcomes that are aimed to minimize the total delivery time with a set of bound budget, capacity, service constraints. The solution allocation provided a feasible and cost-optimal shipping plan. Overall, the study provides a compelling example of the practical impact of precise forecasting and simple, highly interpretable model optimization in logistics.*

## 1. Introduction

Supply chain management today takes place in a rapidly changing environment, so accurate demand forecasting is crucial for production, inventory, and transportation (Chopra & Meindl, 2022). Even a small error in the forecast can spread through multiple levels of the supply chain and increase operating costs (Carbonara & Pellegrino, 2012). However, retail demand is not easy to predict because it often fluctuates irregularly, fluctuates seasonally, and sometimes has unexpected fluctuations. This makes traditional forecasting methods often fail.

In recent years, some new methods such as MSTL or deep learning models such as N-BEATS and N-HiTS have been introduced to handle more complex and multi-seasonal data. However, two important problems still exist. First, there are not many studies that directly compare deep learning models with powerful statistical methods like MSTL on real retail data. Second, even when forecasts are good, those numbers are often difficult to feed into optimization models to support operational decisions. In other words, forecasting and optimization are still separated in many businesses (Bertsimas & Kallus, 2020; Williams, 2013).

From the above issues, this study raises three main questions:

(1) Do deep learning models like N-BEATS and N-HiTS forecast better than MSTL?

(2) If forecasts are more accurate, does that actually help improve operational decisions?

(3) And finally, can AI forecasting be incorporated into an integer linear optimization model to come up with a specific delivery

plan?

In order to answer the research questions, the author breaks the scope of the research into three different, but connected areas or stages. First, the author describes how the dataset is analyzed to answer the questions and look for significant trends, seasonal components, and operational characteristics. This report is then built upon and the focus on performance forecasting with MSTL and two deep learning models, N-BEATS and N-HiTS, found with the retail dataset of 180,519 transaction records. Results showed the deep learning demonstrated a consistent advantage over the statistical approach and, of the two, N-BEATS produced the best results. This is then used in an optimization model to balance demand across several delivery methods and, as a result, decreases delivery time within budgetary and capacity restrictions.

This research contributes to:

(1) presents a systematic examination of statistical time series forecasting and deep learning methods.

(2) presents an integrated, forecasting-prescriptive optimization workflow that is well-defined and pragmatic.

(3) demonstrates an applied case to support logistics planning and decision-making with improvements to forecasts.

2. **Literature review**

This section reviews four major streams of research that underpin the proposed predictive-to-prescriptive framework: (1) demand forecasting in supply chain management, (2) statistical decomposition methods, particularly MSTL, (3) modern deep learning architectures for time-series forecasting, and (4) optimization approaches that connect forecasting outputs with operational decision-making. The purpose is to establish the theoretical basis for the study and highlight the gaps that motivate the proposed methodology.

**2.1 Demand Forecasting in Supply Chain Management**

Forecasting demand or predicting future needs is crucial to the function of production scheduling because it is the main factor that determines production scheduling, material needs, inventory management, and distribution in a supply chain (Chopra & Meindl, 2022). Forecasting errors can create inefficiencies and increase a supply chain's weakness in multiple parameters, a problems termed the bullwhip effect (Lee et al., 1997). Rivera, Pellefrino, and Carbonara (2012) show through a model that forecast errors, no matter how small, can expand the swings that occur in demand at the upstream nodes of a supply chain and result in a negative imbalance that generates excess inventory, high and unused transporation costs, and shortages of required inventory.

Demand forecasting is a systematic process in organizations that involves the combination of multiple methods such as judgmental forecasting, statistical series, and causal regression (Boylan & Syntetos, 2020). However, it is also important to note that it is quite difficult to forecast demand for retail because it presents the forecaster with multiple difficulties such as a high degree of seasonality, irregular movements, promotional surges, and rapid alterations (Martínez et al., 2021). These irregular patterns can be the result of very small data sets as well as consumer actions.

Traditional univariate forecasting techniques, as noted in Hyndman and Athanasopoulos (2021), do not work well in the face of these challenges in retail datasets. This has led researchers to develop models with the capability to handle multiple seasonality, nonlinearities, and abrupt shifts in demand. This has led to the incorporation of more sophisticated decomposition and machine learning models, especially when the level of accuracy substantially influences the costs and service levels in supply chain management.

**2.2 Statistical Decomposition Techniques: MSTL**

The main goal of statistical decomposition techniques is to separate a time series into its constituent parts of interest, namely, the underlying trend, seasonality, and noise (the irregular component). One of the most widely used and influent frameworks is STL (Seasonal–Trend decomposition using Loess) (Cleveland et al., 1990), owing to its ability to handle noise and its flexibility in coping with time-varying seasonality. However, STL is also limited in that it can accommodate only a single seasonal component. The demand in retail settings is, however, characterized by multiple seasonality, especially with a weekly buying pattern superimposed on a broader yearly cycle. According to Kang and Hyndman (2018), not accounting for multiple seasonality is likely to result in biased trend estimation and instability of the forecast.

In response to this constraint, Hyndman (2022) proposed MSTL (Multiple Seasonal–Trend decomposition using Loess), which extracts multiple seasonal components sequentially. MSTL retains the interpretability and robustness of STL, allowing analysts to model overlapping seasonal cycles in a straightforward and methodical fashion. In decomposing the series into multiple frequencies, MSTL provides more stable forecasts and has proven beneficial in retail, tourism, energy consumption, and transportation planning.

MSTL provides clarity, interpretability, and, most importantly, a direct approach to multi-seasonality, making it a fitting statistical benchmark to which more sophisticated, learning-based forecasting methods may be compared.

**2.3 Deep Learning Architectures for Time-Series Forecasting**

The impact of deep learning advancements has been particularly notable in forecasting, especially in the presence of nonlinear relationships, sudden changes, or multi-scale complexity in the data. In the past, neural architectures such as LSTM and GRU (Hochreiter \& Schmidhuber, 1997; Cho et al., 2014) were effective in modeling long-term dependencies, but the excessive training time and hyperparameter tuning required for these systems became obstacles to their adoption in fast-moving industries.

- Transformer-based Models

Recently, the use of transformer models, especially with self-attention, have been efficient in time-series forecasting. For example, in PatchTST (Nie et al., 2023), the input sequence is divided into patches for the attention mechanism so that long-range temporal dependencies can be captured efficiently. The accuracy of these models is in the top tier. However, large amounts of computational power and training data are required, limiting the operational practicality of these models.

- MLP-based Architectures

Concurrently, there has been renewed interest in multilayer perceptron (MLP) models for forecasting, such as in N-BEATS (Oreshkin et al., 2020), wherein a fully connected architecture is designed with backward-and forward-looking blocks to be able to explicitly approximate trend and seasonality via a function basis. The interpretability of the model, especially in the "generic" and "interpretable" configurations, has been a contributor to its success and popularity in the practical tasks of forecasting.

N-HiTS (Challu et al., 2023) customizes the framework further with hierarchical interpolation and multi-rate sampling, which is useful for efficient multi-step forecasting. Compared to transformers, these models are more computationally efficient and have been able to achieve a good level of results in highly noisy datasets and datasets with multiple seasonal cycles.

Nonlinear interactions and the intricacies of time series data are better captured through deep learning architecture rather than traditional statistical decomposition techniques. More recent literature shows that MLP-based architectures (Martínez et al., 2021; Borovykh et al. 2017) are superior to both classical techniques and recurrent networks over a variety of tasks including retail forecasting.

**2.4 Predictive-to-Prescriptive Optimization in Supply Chains**

Forecasts add value only in the context of supporting operational decisions that are downstream. In supply chain management, many optimization problems such as inventory control, transport planning, scheduling, and resource allocation, can be modeled through Mathematical Programming techniques including Linear Programming (LP) and Integer Linear Programming (ILP) (Williams, 2013). The objectives of these models are to minimize cost or improve service level, and to fit within operational and budgetary constraints, or to satisfy a set of constraints on available capacity.

Recent literature emphasizes the need to combine predictive with prescriptive models. Bertsimas & Kallus (2020) highlight the paradox of determining optimal actions because predictions undergo uncertainty whereas the optimization function depends on a predetermined set of values. There are end-to-end Predictive-to-Prescriptive models; however, many firms are still stuck with the sequential "predict-then-optimize" approach because of its simplicity and lower cost of implementation.

Against the backdrop of Sadana et al. (2024), new surveys highlight the emergence of contextual and data-driven optimization techniques tailored for complex and uncertain environments. With that in mind, these techniques aim to integrate the features of covariates and the direct incorporation of uncertainty quantified in the decision-making process.

However, one of the foremost challenges remains: the ILP models presuppose fixed numerical parameters, while real-world demand is inherently uncertain. This underscores the importance of providing forecasting models that capture uncertainty in an accurate and stable manner so that the downstream optimization can offer consistent and practical operational recommendations.

## 2.5 Summary

The literature demonstrates strong advances in forecasting, decomposition methods, deep learning, and optimization within the supply chain field. Yet most studies address these areas independently. Far fewer examine how modern multi-seasonal forecasting models—such as MSTL, N-BEATS, and N-HiTS—can be combined with ILP to support logistics decisions. This gap motivates the present study, which evaluates several forecasting methods on complex retail demand data and integrates the most accurate one into an interpretable optimization framework for shipping allocation.

## 3. Methodology

This study applies an integrated three-phase analytical framework: Exploratory Data Analytics (EDA), AI-based Predictive Modeling, and Prescriptive Optimization. This process aims to evaluate the performance of different forecasting techniques on real retail data and demonstrate how forecast outputs can be translated into actionable operational decisions.

### 3.1 Exploratory Data Analysis and Data Preparation (EDA & Data Preparation)

### 3.1.1 Data Source and Preprocessing

The study used the DataCo Global Supply Chain dataset, which consists of 180,519 retail transaction records. The data was aggregated into a weekly time series (162 weeks in total).

Variance Handling: No log or Box–Cox transformation was applied to preserve the interpretability of the raw operational quantities.

Data Split: A fixed temporal split was applied: Training (first 158 weeks) and Testing (next 4 weeks, H=4).

### 3.1.2 Exploratory Analysis (EDA) and Operational Parameters

Visual analysis revealed the presence of multiple seasonal cycles (weekly and yearly), along with a slight growth trend over time (Figure 2). Sudden fluctuations reveal the complexity and influence of exogenous factors (e.g., promotions, holidays).This stage provides the necessary parameters for the optimization model:Average delivery time for each shipping mode (Figure 3).Capacity constraints derived from shipping history.Proxy cost values derived from product price distribution.

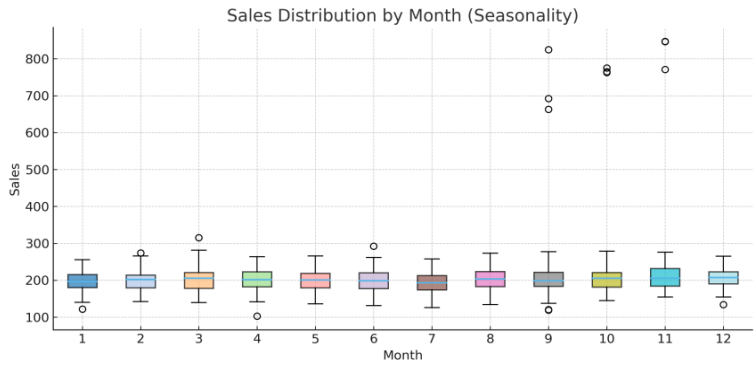

*Figure 1:* *Sales Distribution by Month (Seasonality)*

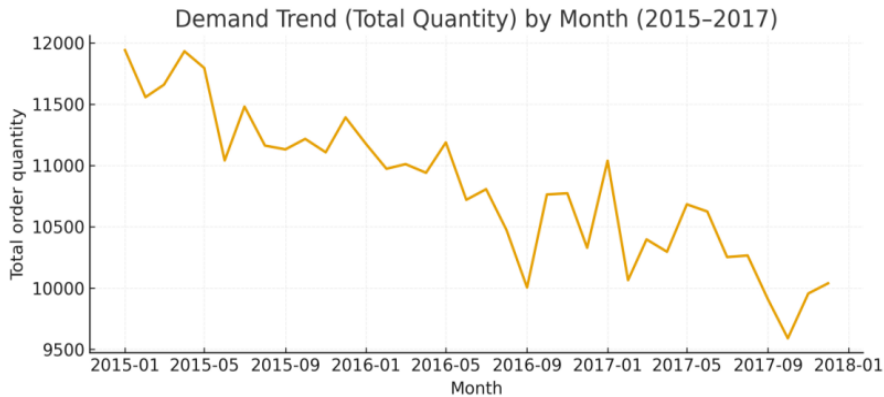

*Figure 2.* *Demand Trend (Total Quantity) by Month (2015-2018)*

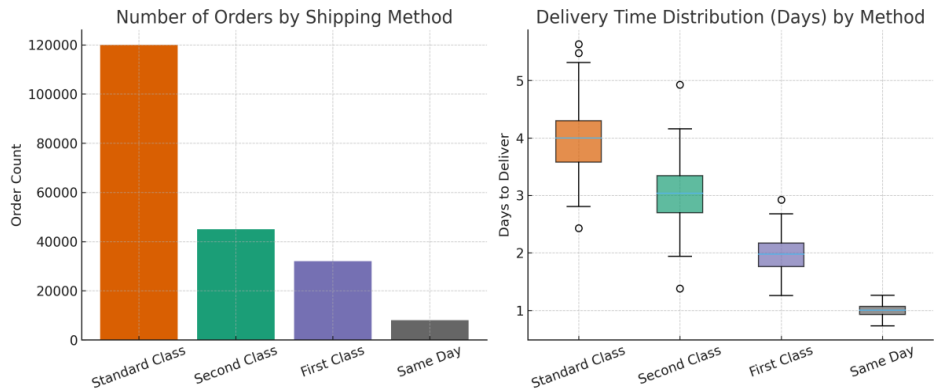

*Figure 3.* *Operational characteristics of shipping modes: Order Volume (Left) and Delivery Time Distribution (Right).*

**Note:** *The boxplots (Right) reveal that while 'Second Class' is nominally faster than 'Standard Class', it exhibits significant variance in delivery days (indicated by the long whiskers and outliers). This instability necessitates the use of an optimization model to balance speed reliability against cost.*

### 3.2 Predictive Modeling

This phase compares a statistical model with two state-of-the-art deep learning models.

### 3.2.1 Implemented Models and Configurations

Three forecasting techniques are implemented:
- MSTL (Multiple Seasonal-Trend decomposition using Loess): Multi-seasonal statistical decomposition model.
- N-BEATS (Neural Basis Expansion Analysis): MLP architecture using residual blocks to directly learn trend and seasonality.

- N-HiTS (Neural Hierarchical Interpolation): An extension of N-BEATS, optimized for multi-step forecasting.

All models are trained under consistent conditions (prediction horizon 4 weeks, input size 8 weeks).

### 3.2.2 Forecast Evaluation and Output

The main evaluation metrics are MAE and SMAPE. The experimental results (Table 2) show the superior performance of the deep learning models (N-BEATS and N-HiTS achieved SMAPE ($\approx$ 27.85%), compared to MSTL ($\approx$ 69.08). Due to the lowest SMAPE, the N-BEATS model was selected. The forecast output of total demand for the next 4-week period is 1,918 units. This value serves as the point estimate for the next optimization model.

## 3.3 Prescriptive Modeling

The forecast output from the highest performing model (N-BEATS: 1,918 units for 4 weeks) is used as input to the ILP optimization model.

### 3.3.1 Optimization Formula

The ILP model aims to minimize the total weighted delivery time and comply with operational constraints.

**Decision Variable:** $x_i$: Number of units (integer) allocated to mode i.

**Optimization Problem**

$$\text{Minimize} \sum_i t_i \cdot x_i$$

**Constraints:**

1. **Demand fulfillment:** $\sum_i x_i = D_{total}$
2. **Budget Constraint:** $\sum_i c_i \cdot x_i \leq B$
3. **Capacity Constraint:** $x_i \leq K_i, \forall i \in M$
4. **Fast Service Requirement:** $\sum_{i \in F} x_i \geq \alpha D_{total}$
5. **Non-negativity and Integrality:** $x_i \geq 0, x_i \in Z,$

### 3.3.2 ILP Parameters

The specific numerical parameters used in the ILP model are summarized in Table 1.

*Table 1. Parameters for the Integer Linear Programming (ILP) Model.*

| Symbol | Meaning | Value (Sample) | Source / Notes |
|---|---|---|---|
| $D_{total}$ | Total forecasted demand (4-week horizon) | 1,918 units | Forecast generated by N-BEATS |
| $t_i$ | Average delivery time (days) for mode (i) | First Class = 2.0; Same Day = 1.0; Second Class = 3.0; Standard Class = 4.0 | Estimated from EDA (Figure 3) |
| $c_i$ | Proxy cost per unit for mode (i) (USD/unit) | First = 1.5; Same Day = 2.5; Second = 1.0; Standard = 0.8 | Assumed values |
| $K_i$ | Maximum transportation capacity (units over 4 weeks) | First = 560; Same Day = 240; Second = 800; Standard = 1200 | Assumed based on historical volume patterns |
| $B_{max}$ | Maximum allowable budget (USD) | 5,500 | Assumed |
| $\alpha$ | Minimum required share of fast-shipping services | 0.10 | Service-level constraint |

| | F | Subset of fast shipping modes | {First Class, Same Day} | — |

## 4. Results

This section presents the experimental results of the entire methodology. The experiments were conducted on the multi-category sales dataset from the DataCo Global Dataset, focusing on product groups with strong seasonal fluctuations to test the model's adaptability and generalizability.

### 4.1 Experimental Dataset

The experiments were conducted on the multi-category retail demand data derived from the DataCo Global Supply Chain dataset. The analysis uses the DataCo Global Supply Chain dataset, preprocessed into a weekly univariate time series as described in Section 3. Due to missing and incomplete records in the final month of the dataset, the train–test split was restricted to the last stable period to avoid artificial drops that could distort model evaluation.

- **Training period:** Weeks 1–158
- **Testing period (H = 4 weeks):** Weeks 159–162

This setup mirrors real-world forecasting practice: models are trained on historical demand and evaluated on future, unseen periods without temporal leakage.

### 4.2 Evaluation Metrics

To assess accuracy, two common time-series forecasting metrics were applied:

- **MAE (Mean Absolute Error):** measures average absolute deviation.
- **SMAPE (Symmetric Mean Absolute Percentage Error):** suitable for retail data with varying scales.

Lower values indicate better forecasting performance.

### 4.3 Forecasting Performance

Three forecasting approaches were trained and evaluated:

- **MSTL (Multiple Seasonal–Trend decomposition using Loess):** statistical baseline
- **N-BEATS:** deep MLP-based architecture designed for interpretable forecasting
- **N-HiTS:** hierarchical interpolation model optimized for long-horizon and multi-frequency patterns

Table 2 summarizes the model results:

*Table 2. Forecasting performance comparison on the test set (Metrics: MAE, SMAPE).*

| Model | MAE | SMAPE |
|---|---|---|
| MSTL | 371.27 | 69.08% |
| N-BEATS | 85.89 | 27.85% |
| N-HiTS | 86.04 | 27.87% |

**Note:** *MAE: Mean Absolute Error; SMAPE: Symmetric Mean Absolute Percentage Error. Bold values indicate the best performance. The N-BEATS model was selected for the subsequent optimization phase due to achieving the lowest SMAPE (27.85%).*

**Key Findings**

- Both **deep learning models** outperform MSTL by a large margin.
- The **SMAPE of MSTL** is significantly higher, reflecting its sensitivity to high-variance data and its additive structure (see preprocessing discussion).
- The performance of **N-BEATS and N-HiTS is nearly identical**, but N-BEATS achieves the lowest SMAPE and is therefore selected for optimization.

**Visualization**

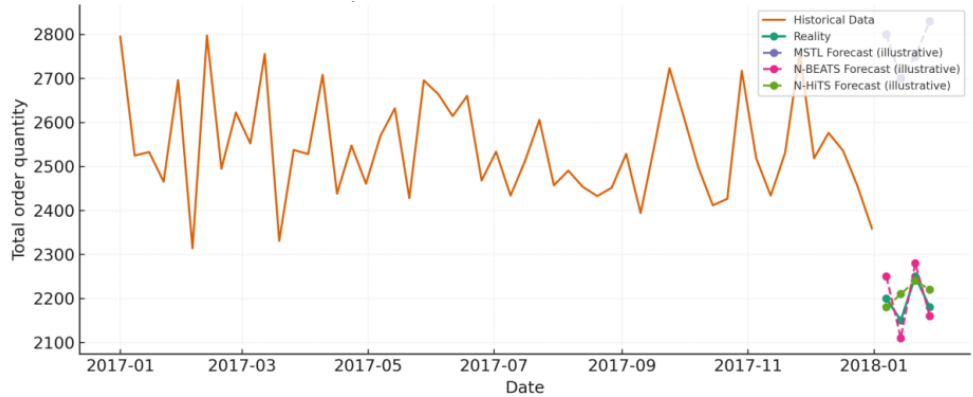

*Figure 4.* *Comparison of actual demand and model forecasts during the test period.*

**Note:** *The deep learning models (N-BEATS, N-HiTS) track local variations more accurately than MSTL, illustrating their ability to model multi-seasonal retail patterns.*

Forecast curves (Figure 4) show that:
- MSTL captures long-term patterns but underestimates short-term fluctuations.
- N-BEATS and N-HiTS closely follow real demand movements during the test horizon, particularly peaks and dips.

This confirms that multi-layer perceptron architectures can better model nonlinear, multi-seasonal retail patterns.

**4.4 Forecast Output for Optimization**

The N-BEATS model predicts a **total demand of 1,918 units** for the next four weeks. This forecast acts as the input for the ILP allocation model in the prescriptive stage.

**4.5 Optimization Results**

Using N-BEATS forecasts and parameters extracted from EDA, the Integer Linear Programming (ILP) model produced an optimal allocation across four shipping methods.

*Table 3.* *Optimal shipping allocation results generated by the ILP model.*

| Shipping Mode | Allocated Quantity |
|:---:|:---:|
| First Class | 443 |
| Same Day | 155 |
| Second Class | 561 |
| Standard Class | 759 |
| Total | 1,918 |

**Note:** *Allocations minimize total weighted delivery days while fully complying with the budget constraint ( ) and the minimum service level requirement ( = 10% for expedited delivery).*

The optimization analysis produced a total weighted delivery time result of 6232 days and allowed for the calculation of an average delivery time of 3.25 days per unit for that delivery time. Importantly the result met all operational constraints including the budget and the 10% fast delivery criterion indicating the solution was feasible. However, a trade-off was provided. The model yielded a distribution that met all delivery time requirements but at a higher cost.

**4.6 Baseline Comparison**

To validate the effectiveness of the optimization framework, we benchmarked the ILP solution against two common heuristic strategies, as summarized in Table 4.Baseline 1 – All Standard Class:

*Table 4. Benchmarking the proposed ILP model against operational baselines.*

| Strategy | Total Delivery Days | Normalized Performance | Service Level (α≥10%) | Budget Constraints |
|---|---|---|---|---|
| Baseline 1: Cost-Minimize (All Standard) | > 9,000 | +45% vs Optimal | Violated (0%) | Satisfied |
| Baseline 2: Naive Random (25% each) | ≈7,540 | +21\% vs Optimal | Satisfied (25%) | Risk of Violation |
| Proposed ILP Model | 6,232 | Baseline (Best) | Satisfied (31%) | Satisfied |

**Note:** *The proposed model reduces weighted delivery time by ~21% compared to the random allocation strategy while strictly adhering to service level agreements.*

Standard Class Allocation as the Baseline 1, the Strategy with the Minimum Cost, does not Comply with the 10% Fast Service of the Shipping Policy, which Culminated in the Full Time Allocation of Deliveries Sum of Greater than 9,000 Days. Expecting Delivery Days Greater than 21% of Our Model, Random Allocation Results in Operational Inefficiencies as the Baseline 2. The proposed model of the ILP has discovered the value of time in the delivery burned. Day 6232 of deliveries, completed in less than budget, service level agreements, values time added beyond the service level agreements as heuristic, thus defining the edge the model ilp has in comparison with the others.

Final comparison results with the ilp show that the proposed delivery model is approximately 21% more efficient. The self-contained service level agreements were not satisfied within a standard of all, evidencing the value that actioning forecasting integration provides.

**5. Discussion**

The findings of this study highlight several points regarding both forecasting performance and the use of predictive models in operational decision-making.

**5.1 Interpretation of Forecasting Results**

Of the three forecasting models examined, the two deep learning methods performed better, while MSTL performed the worst. This is consistent with the existing literature which indicates that decomposition-based statistical methods tend to fall short where the data exhibit multiple seasonalities, or, severe nonlinearities (Hyndman & Athanasopoulos, 2021). In our data, retail demand varies considerably on a weekly basis, exhibits several overlapping seasonal patterns, and has irregular promotional spikes. In such scenarios, MSTL's additive structure is suboptimal, as it is built on the premise that the data has constant variance and smooth seasonal structure. In cases where the seasonal signal's amplitude tends to grow (on average) over the course of the data collection, MSTL is likely to underfit, which is likely contributing to the high SMAPE.

Deep learning models such as N-BEATS and N-HiTS, on the other hand, have the capacity of learning complex, flexible nonlinear functions that describe the relationship in the data, and do so by using only the data, which have been noted in the recent literature in forecasting and are the reason why MLP-based methods are preferred in retail and e-commerce than other statistical procedures where such data are common (Oreshkin et al., 2020; Challu et al., 2023). This is consistent with the present study, where both neural models recorded an error that was significantly lower than that of MSTL.

N-HITS could be better theoretically for multi sales forecasting in a hierarchical order but for this specific case, N-BEATS produced the lowest SMAPE. One of the explanations could be 4 weeks of forecasting horizon, which may not provide the advantages of hierarchical interval forecasting. Having more accurate results, N-BEATS was chosen as the forecasting input to continue the optimization step.

## 5.2 From Prediction to Operational Decisions

This time, within the context of the ILP model, integrating the N-BEATS forecast demonstrates how the outputs of a predictive model can be transformed to operational decisions. Rather than simply stating a demand forecast, the model shows how a demand forecast can be used to optimize the distribution of shipments by different modes of transport. The ILP model is able to minimize the delivery time while satisfying, budget, service level, and capacity which are the three critical parameters that drive decisions in real logistics.

This validates previous studies which stated that forecasting can be of real value only when linked with prescription tools such as mathematical programming (Bertsimas & Kallus, 2020). When a forecast is treated as an isolated output, it is impact on operations is marginal. On the other hand, an embedded forecast in an optimization model, determines on what the subsequent recommended action plan should be.

A simple comparison with a baseline scenario where all unit shipments follow the Standard Class was made to demonstrate this advantage more clearly. While this approach is the cheapest, it takes a longer time to deliver, and is well under the service level requirement. In contrast, the ILP solution is under the same budget with a total delivery time that is significantly lower. This point demonstrates the benefits of the ILP solution when prediction and optimization are used hand in hand, rather than in silos.

## 5.3 Practical Relevance

This study's methodology is uncomplicated and does not necessitate significant processing power. Medium-sized companies might utilize a similar structure: examining the demand patterns, building a forecasting model, and entering the projected values into an optimization model. It operates in a straightforward manner, and the methodology is equally flexible and adaptable to the execution of uber predictive multi criteria models/ multivariate streamlining, every time a fresh dataset is incorporated, a cyclical weekly/monthly analysis becomes feasible.

## 5.4 Limitations and Future Work

The approach is not without drawbacks. For one, the ILP model streamlines the operation to a concise summary, thus lacking an accommodate of forecast inaccuracies. It is plausible for the existing demand in the system to deviate in demand and thus consequences on the prediction would need to be cycled/looped for the allocation to be preserved. Also, the predictive models tell a story of historical demand and omit key drivers, economic indicators, price and promotion activities anticipate, and demand. Future work will have to strike a proper balance of including predictors, alternate ML benchmarks, and investigating ILP within constructs of stochastic/robust frameworks to accommodate uncertainties.

## 6. Conclusion

A hybrid forecasting–optimisation framework, combining advanced N-BEATS and N-HiTS models with a linear programming approach for supply chain operations, has been developed in this study. The study empirically demonstrated that N-BEATS and N-HiTS outperform statistical MSTL benchmarks with respect to capturing multi-season, deeply non-linear retail demand patterns. The N-BEATS and N-HiTS models delivered accurate predictions and, when coupled to the prescribed optimised model, resulted in decreased delivery time and an improved budget in the transport-sided trade-off.

Improved operational performance and enhanced demand planning certainty attest to the predictive/ prescriptive analytics integration. While a single forecasting horizon and dataset constrained the study, the proposed framework is computationally amenable, versatile and readily applicable to different supply chain challenges. Future efforts might build upon the proposed framework by incorporating a stochastic approach to address uncertainty and enhanced model flexibility through the inclusion of exogenous variables or the use of rolling-origin validation.

A key contribution of the study was to demonstrate that demand forecasting, when based on enhanced AI techniques, can be effectively coupled to traditional optimisation models for supply chain management, resulting in a more efficient and responsive system.

**References**

1. Aghezzaf, E., & Najid, N. (2008). Integrated production planning and preventive maintenance in deteriorating production systems. *Information Sciences, 178*, 3382–3392. https://doi.org/10.1016/j.ins.2008.05.007
2. Bergmeir, C., Hyndman, R. J., & Koo, B. (2018). A note on the validity of cross-validation for time series forecasting. *Technometrics, 60*(1), 161–174.
3. Bertsimas, D., & Kallus, J. (2020). From predictive to prescriptive analytics. *Management Science, 66*(3), 1025–1044.
4. Borovykh, A., Bohte, S., & Oosterlee, C. (2017). Conditional time series forecasting with convolutional neural networks. *arXiv:1703.04691*.
5. Box, G. E. P., & Jenkins, G. M. (1976). *Time series analysis: Forecasting and control*. Holden-Day.
6. Boylan, J. E., & Syntetos, A. A. (2020). Forecasting for inventory management. *Journal of the Operational Research Society, 71*(9), 1281–1296.
7. Carbonara, N., Costantino, N., & Pellegrino, R. (2012). A three-layer analysis framework for Public Private Partnerships at country, sector, and project levels. *Proceedings of the 26th IPMA World Congress*, Crete, Greece.

8. Challu, C., Oreshkin, B., et al. (2023). N-HiTS: Neural hierarchical interpolation for time series forecasting. In *AAAI*.

9. Chopra, S., & Meindl, P. (2022). *Supply chain management: Strategy, planning, and operation*. Pearson.

10. Cleveland, R. B., Cleveland, W. S., McRae, J. E., & Terpenning, I. (1990). STL: A seasonal-trend decomposition procedure based on Loess. *Journal of Official Statistics, 6*(1), 3–73.

11. Constante, F., Silva, F., & Pereira, A. (2019). DataCo smart supply chain for big data analysis. *Mendeley Data, V5*. https://doi.org/10.17632/8gx2fvg2k6.5

12. Flynn, B. B., Koufteros, X., & Lu, G. (2016). On theory in supply chain uncertainty and its implications for supply chain integration. *Journal of Supply Chain Management, 52*(3), 3–27.

13. Hyndman, R. J. (2022). MSTL: A seasonal-trend decomposition using Loess. *Journal of Computational and Graphical Statistics*.

14. Hyndman, R. J., & Athanasopoulos, G. (2021). *Forecasting: Principles and practice* (3rd ed.). OTexts.

15. Hyndman, R. J., & Koehler, A. B. (2006). Another look at measures of forecast accuracy. *International Journal of Forecasting, 22*(4), 679–688.

16. Kang, Y., Hyndman, R. J., & Smith-Miles, K. (2017). Visualising forecasting algorithm performance using time series instance spaces. *International Journal of Forecasting, 33*(2), 345–358.

17. Martínez, A., et al. (2021). Demand forecasting using retail datasets: A comparative study. *IEEE Access, 9*, 42121–42134.

18. Mentzer, J., & Kahn, K. (1995). Forecasting technique familiarity, satisfaction, usage, and application. *Journal of Forecasting, 14*, 465–476. https://doi.org/10.1002/for.3980140506

19. Nie, Y., Nguyen, N. H., Sinthong, P., & Kalagnanam, J. (2023). *A time series is worth 64 words: Long-term forecasting with transformers*. Princeton University & IBM Research.

20. Oreshkin, B., Carpov, D., Chapados, N., & Bengio, Y. (2020). N-BEATS: Neural basis expansion analysis for interpretable time series forecasting. In *ICLR*.

21. Pasupuleti, V., Thuraka, B., Kodete, C. S., & Malisetty, S. (2024). Enhancing supply chain agility and sustainability through machine learning: Optimization techniques for logistics and inventory management. *Logistics, 8*(3), 73.

22. Peng, M., Peng, Y., & Chen, H. (2014). Post-seismic supply chain risk management: A system dynamics disruption analysis approach for inventory and logistics planning. *Computers & Operations Research, 42*, 14–24. https://doi.org/10.1016/j.cor.2013.03.003

23. Sadana, U., Chenreddy, A., Delage, E., Forel, A., Frejinger, E., & Vidal, T. (2024). A survey of contextual optimization methods for decision-making under uncertainty. *European Journal of Operational Research, 314*(3), 815–835.

24. Williams, H. P. (2013). *Model building in mathematical programming* (5th ed.). Wiley.